\documentclass[10pt,twocolumn,letterpaper]{article}
\usepackage{cvpr}
\usepackage{times}
\usepackage{epsfig}
\usepackage{graphicx}
\usepackage{amsmath}
\usepackage{amssymb}
\usepackage[ruled]{algorithm2e}
\usepackage{bm}
\usepackage{multirow}
\usepackage{float}
\usepackage{diagbox,cite,cases,dsfont,mathrsfs,booktabs,multirow,psfrag,subfigure,stfloats,amsthm}

\usepackage[pagebackref=true,breaklinks=true,letterpaper=true,colorlinks,bookmarks=false]{hyperref}

\cvprfinalcopy 

\begin{document}

\title{Domain Adaptive Person Re-Identification via Camera Style Generation and Label Propagation}
\author{Chuan-Xian Ren\thanks{C.X. Ren and B.H. Liang are with the School of Mathematics, Sun Yat-Sen University, Guangzhou 510275, China. Z. Lei is with the Center for Biometrics and Security Research and National Laboratory of Pattern Recognition, Institute of Automation, Chinese Academy of Sciences, Beijing 100190, China.}, \:\: Bo-Hua Liang, \:\: Zhen Lei
}
\maketitle

\begin{abstract}
Unsupervised domain adaptation in person re-identification resorts to labeled source data to promote the model training on target domain, facing the dilemmas caused by large domain shift and large camera variations. The non-overlapping labels challenge that source domain and target domain have entirely different persons further increases the re-identification difficulty. In this paper, we propose a novel algorithm to narrow such domain gaps. We derive a camera style adaptation framework to learn the style-based mappings between different camera views, \textit{from the target domain to the source domain}, and then we can transfer the identity-based distribution from the source domain to the target domain on the camera level. To overcome the non-overlapping labels challenge and guide the person re-identification model to narrow the gap further, an efficient and effective soft-labeling method is proposed to mine the intrinsic local structure of the target domain through building the connection between GAN-translated source domain and the target domain. Experiment results conducted on real benchmark datasets indicate that our method gets state-of-the-art results.
\end{abstract}

\section{Introduction}\label{sec:introduction}

Person re-identification (person re-ID) aims to find the same person among a camera network. Its applications in security and surveillance draw attention from both academia and industry on person re-ID and thus impel the development of related algorithms. Particularly, supervised methods for person re-ID produce good results in the literature~\cite{sun2017beyond, zhang2017alignedreid}.

\begin{figure}[htbp]
\centering{{\includegraphics[width=0.46\textwidth]{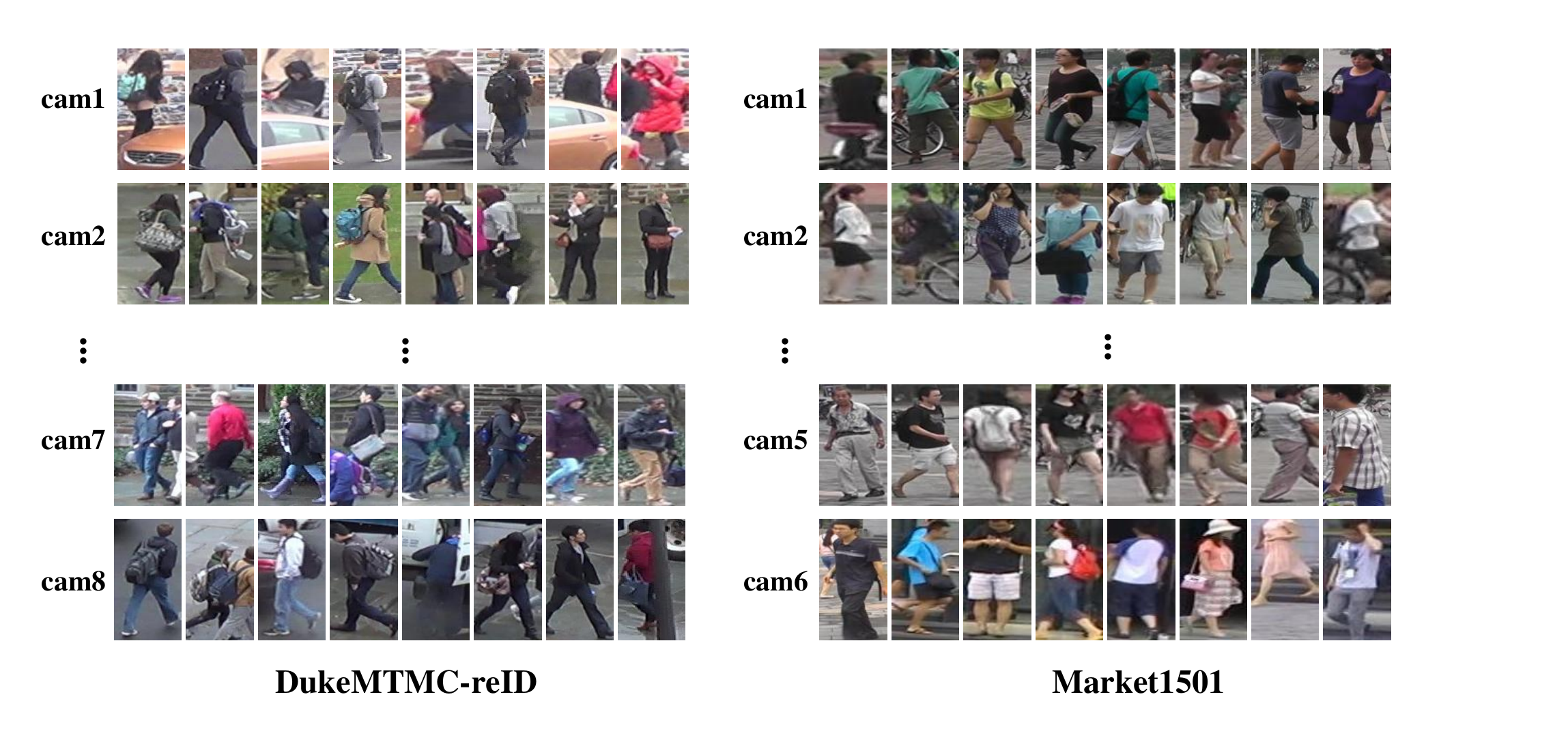}}\caption{Exemplar samples from Market1501 and DukeMTMC-reID with their camera labels specified.}\label{fig:camera variations}} 
\end{figure}

However, in many scenarios labeled data are unavailable in an interested target domain. Labeling target domain in a manual manner is extremely expensive on account of correctly finding out the same person among different cameras, even in a network scene with moderate amount of cameras. Thus supervised learning in target domain is infeasible in such case. Previous works~\cite{ma2014covariance, matsukawa2016hierarchical, liao2015person, zheng2015scalable} attempt to learn re-ID models in a completely unsupervised way through extracting hand-crafted features. But label information is important for the system to learn discriminative features. One case is that we can adapt the system trained on the labeled data from other domains, which are called source domains. Recent works~\cite{Fan01Unsupervised, wei2017person} test the trained-ready systems on the target domain but just obtain bad performance. That is, re-ID systems are sensitive to the domain gaps between different scenarios, which may due to cameras discrepancy on lighting conditions, resolutions, human race, seasons, backgrounds, etc. Some examples are shown in Fig.~\ref{fig:camera variations}. The camera styles of Market1501~\cite{zheng2015scalable} and DukeMTMC-reID~\cite{ristani2016performance} are quite different. For example, due to seasons and viewpoints, clothing style between the two domains exhibits large discrepancy, and the lighting variation of DukeMTMC-reID is larger than that of Market1501. Additionally, images in DukeMTMC-reID have more distinct backgrounds than Market1501. This indicates that the model trained on the source domain may have weak generalization ability on the target domain. It brings a new problem that how can we adapt the model to a target domain that we are interested in. This refers to an unsupervised learning setting which attempts to use some valuable information in existing labeled data from source domains, i.e, unsupervised domain adaptation (UDA) in person re-ID, in which data from the source domain is fully labeled while data from the target domain are unlabeled.

\begin{figure*}[htbp]
\centering{{\includegraphics[width=0.95\textwidth]{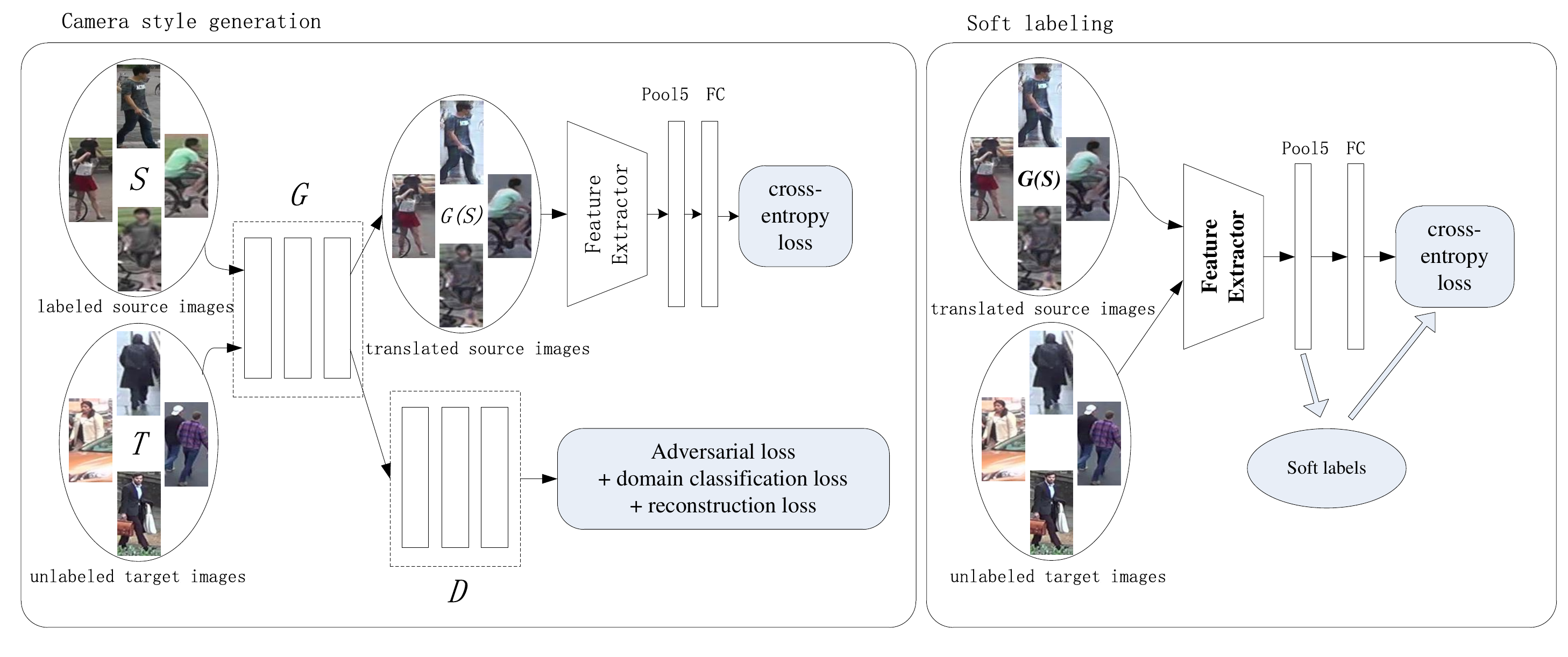}}\caption{Flowchart of the CSGLP algorithm. The left diagram is camera style adaptation, in which $G$ and $D$ is the generator and discriminator of StarGAN, respectively. $G$ makes efforts to translate the camera styles from $\mathcal{T}$ to $\mathcal{S}$. The right diagram corresponds to the KNN-based soft-labeling module.}\label{fig:framework}} 
\end{figure*}

Two main challenges exist in such a setting. One is the camera variations challenge that both between-domain and within-domain camera style variations are large (see Fig.~\ref{fig:camera variations}), which may cause failure to re-ID systems trained on the source domain only when directly testing on a target domain, as the model cannot learn target camera-invariant features. Another is the non-overlapping labels challenge that pedestrians in source and target domains are totally different, making the task more difficult.

Several methods have made efforts to tackle such problems. For the camera variations challenge, refs.~\cite{Deng01Image, wei2017person} utilize generative adversarial networks (GAN)~\cite{goodfellow2014generative} to facilitate the learning task. GAN, typically consisting of a generator and a discriminator, learns mappings from one distribution to another with unlabeled even random data. Extensive variants~\cite{mirza2014conditional, zhu2017unpaired, kim2017learning, yi2017dualgan, liu2017unsupervised, Choi01StarGAN} have shown remarkable performance on image-to-image translation tasks. Thus in unsupervised domain adaptation re-ID, following CycleGAN~\cite{zhu2017unpaired}, Similarity Preserving cycle-consistent Generative Adversarial Network~\cite{Deng01Image} (SPGAN) and Person Transfer Generative Adversarial Network (PTGAN)~\cite{wei2017person} propose image-to-image translation frameworks for re-ID to relieve the camera gaps between source and target domains. Both of them attempt to train a discriminative model on the GAN-translated data which originally come from the source domain. Such translated images share the same camera styles with the target domain while preserving pedestrian labels. However, these methods either transfer the global style of the source domain to the target domain or train multiple CycleGANs to learn translation mappings between specific camera styles. They are unable to capture
camera variations within one domain in an efficient way. This is time-consuming in that multiple GANs need to be trained, and also data-deficit in that only a subset of the data is used to train each mapping.

For the non-overlapping labels challenge, refs.~\cite{ye2017dynamic, liu2017stepwise, Fan01Unsupervised} stress importance on reliable label estimation for the target domain. While UMDL \cite{peng2016unsupervised} proposes a multi-task dictionary learning method combining the source and target data to obtain a discriminative target-domain representation. Nevertheless, these methods concentrate on only one aspect of these two challenges, and specially camera variations in the target domain have not got enough attention.

In this paper, we propose a new algorithm framework to tackle these two challenges together. Fig.~\ref{fig:framework} shows our motivation and algorithm flowchart. For simplicity, we abbreviate our camera style generation and label propagation method to CSGLP hereinafter. To learn the transfer mappings between different camera views and multiple datasets/domains simultaneously, we explore StarGAN~\cite{Choi01StarGAN} as camera style adaptation network. This framework achieves multiple style adaptations across cameras by using a unique generative adversarial network. For the non-overlapping labels challenge, based on the fact that the GAN-translated source domain has similar data distribution with the target domain, we propose a KNN-based method to generate soft labels for the target domain. The soft labels partly reflect local structure information of the unlabeled data. The cross entropy loss is then employed to guide the training process of the person re-ID model. This can further improve the person re-ID accuracy on the target domain.

Owe to that camera labels are easy to obtain, in this paper we assume camera labels are available for both source domain and target domain. Our main contributions are summarized as follows.
\begin{enumerate}
\item A new method, i.e., CSGLP, is proposed to narrow the domain gap between the source domain and the target domain and generate discriminative representations for the target domain. In particular, the style adaptation is simultaneous cross-domain and cross-camera for person re-ID.
\item StarGAN is used to transfer camera styles from the target domain to the source domain. To the best of our knowledge, this utilization is novel in the literature. By introducing the camera labels and the domain labels, StarGAN generates the new features with only one generator and one discriminator.
\item To further narrow the distribution gap between different domains and predict identities of the target domain images, a soft labeling method is constructed on the connections between the translated source domain and the target domain. Experiment results show competitive performance when compared with some state-of-the-art methods.
\end{enumerate}

The rest of this paper is organized as follows. Section~\ref{sect:related-work} briefly reviews some related work about current achievements in the literature. Section~\ref{sect:CSGLP} proposes our CSGLP algorithm for unsupervised domain adaptation in person re-identification. Experiment results and analysis are presented in Section~\ref{sect:experiments}, in which several state-of-the-art methods are compared with CSGLP. Section~\ref{sect:conclusion} concludes the paper and shows future work.

\section{Related Work}~\label{sect:related-work}
In this section, we have a brief review on some related work including image-to-image translation, unsupervised domain adaptation and unsupervised person re-ID.

\textbf{Image-to-Image Translation}. Image-to-image translation aims to translate an image to another one with given attributes changed. Recent literature based on GANs~\cite{goodfellow2014generative} has shown impressive results in image-to-image translation. Typically, GANs consist of a generator $G$ and a discriminator $D$, aiming to learn the true data distribution by a min-max game. Let $x$ be an image (usually in the tensor form with three channels) sampled from the given dataset, and $z$ be a random vector which obeys Gaussian or other distribution. $p_x$ and $p_z$ are corresponding probability distribution functions. The generator $G$ tries to generate fake images, such as $G(z)$, to fool the discriminator $D$, while the discriminator tries to classify the real images and the fake images. It is essentially a generative framework in which the discriminator $D$ is introduced to find against with the generator $G$ a min-max game as below.
\begin{equation}
\label{eq:gp}
\min_{G}\max_{D} \mathds{E}_{x \in p_{x}}[\log D(x)] + \mathds{E}_{z \in p_{z}}[\log (1-D(G(z)))].
\end{equation}
Because $\log (1-D(G(z)))$ and $-\log (D(G(x)))$ have the same optimization direction, the latter is often used for the sake of stability. Blurry images will not be tolerated since they look obviously fake, thus finally the generator can learn data distribution of real images and generated images that look exactly like the real ones.

A stream of relevant methods are proposed to improve the learning capacity of GANs. cGANs\cite{mirza2014conditional} and its variant \cite{isola2017image} learn generators by combining the original adversarial loss with a $\ell_1$ loss which force the generated images to be near the ground truth output under the $\ell_1$ distance. However, they need paired data constrains for scalability. Thus unpaired image-to-image frameworks \cite{zhu2017unpaired, kim2017learning, yi2017dualgan, liu2017unsupervised} have been proposed to alleviate this limitation. In \cite{zhu2017unpaired, kim2017learning, yi2017dualgan}, a cycle consistency loss is introduced to preserve the image contents and only change the domain-related parts. However, in these frameworks, one model needs to be trained for every domain pair mapping at a time. This cannot meet the scalability in handling multiple domains. StarGAN \cite{Choi01StarGAN} tackles this problem by introducing an auxiliary classifier \cite{odena2016conditional} to allow the discriminator of GAN to control multiple domains. Iterative training approaches that alternates between multiple domains make it possible that the generator learns multiple mappings simultaneously.

\textbf{Unsupervised domain adaptation}. The setting of unsupervised domain adaptation (UDA) is that source data is labeled while the target data are unlabeled, which is consistent to our setting in this paper. Among those UDA methods, there are two main streams: one attempts to find a domain-invariant feature space for both source domain and target domain\cite{long2015learning, ganin2016domain} and the other learns a mapping between source domain and target domain\cite{gong2012geodesic, fernando2013unsupervised, sun2016return, hoffman2017cycada, bousmalis2017unsupervised, liu2016coupled}. However, many of these methods are based on the setting that the same labels are shared by source domain and target domain, while in this paper, we have completely different identities from the source domain to the target domain. In other words, directly applying these UDA methods to our setting is impractical and infeasible.

\textbf{Unsupervised Person Re-ID}. While supervised person re-ID methods have achieved high accuracies due to the access of deep learning algorithms and large scale datasets, the great demand on labeled data limits their generalization and applications. Unsupervised methods avoid expensive artificial data labeling or annotation. One typical type of unsupervised methods is to extract hand-crafted features \cite{ma2014covariance, matsukawa2016hierarchical, liao2015person, zheng2015scalable} without learning. It is straightforward. But such methods may loss valuable information in labeled data from external domains, which can be exploited to obtain discriminative features for UDA tasks. UMDL \cite{peng2016unsupervised} resorts to the dictionary learning approach to obtain a dataset-shared but target data-biased representation with the labeled source domain. SPGAN\cite{Deng01Image} and PTGAN\cite{wei2017person} use the similar settings with UMDL to learn image translation for unsupervised person re-ID. Specifically, SPGAN uses an additional SiaNet to preserve the ID-related information, while PTGAN uses an extra PSPNet\cite{zhao2017pyramid} to make person ID be ignored by the generator. Both of them cannot efficiently capture the camera variations with one generator.

\begin{figure}[htbp]
	\centering{{\includegraphics[width=0.48\textwidth]{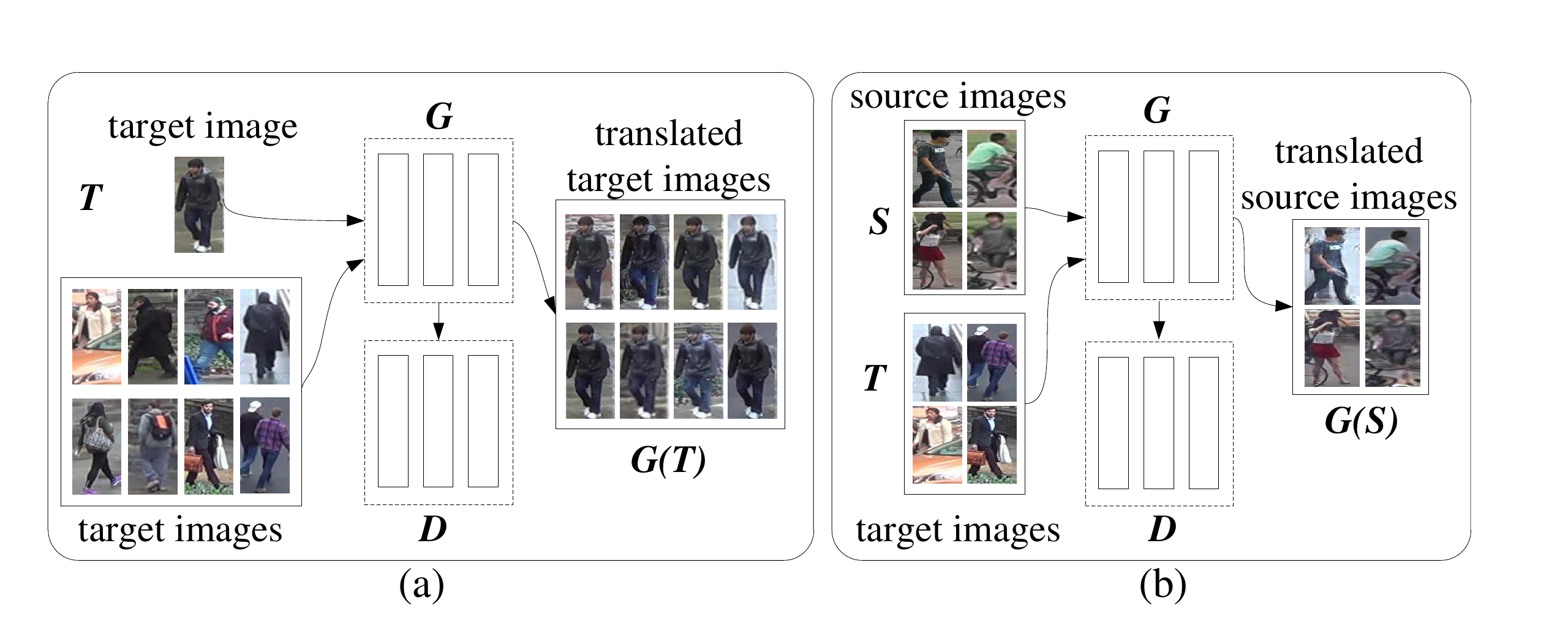}}\caption{Schedule difference between HHL~\cite{Zhong2018retri} and CSGLP. (a) The camera style is adapted between images inside the target domain only; (b) The camera style is transferred from the target domain to the source domain.}\label{fig:method-diff}}
\end{figure}

We note that Zhong et al. propose a Hetero-Homogeneous Learning (HHL) method~\cite{Zhong2018retri} to address the domain adaptive person re-ID problems, and the StarGAN approach is also used for camera style adaptation. However, it has significant difference between HHL and CSGLP. We show their schedule flowcharts of camera style translation in Fig.~\ref{fig:method-diff}. HHL considers style translation between images inside the target domain only, while CSGLP makes efforts to transfer camera styles from the target domain to the source domain. The cross-domain style translator is expected to play an active role in distribution approximation and feature matching.

To alleviate the view-specific interference in person re-ID tasks, Ref.~\cite{yu2017cross} learns one projection for a camera to project camera variations into a shared space. There are several methods~\cite{ye2017dynamic, liu2017stepwise, Fan01Unsupervised} on label estimation for unlabeled target domain. Ye et al.~\cite{ye2017dynamic} and Liu et al.~\cite{liu2017stepwise} leverage cross-camera labeling association for label estimation. Progressive Unsupervised Learning (PUL)~\cite{Fan01Unsupervised} alternates clustering and fine-tuning to progressively train a discriminative model. Those methods including \cite{yu2017cross, ye2017dynamic, liu2017stepwise, Fan01Unsupervised} are all based on the original labeled source domain and the unlabeled target domain, while our CSGLP method can effectively use the relationship between translated labeled data from the source domain and unlabeled data of the target domain.

\section{The CSGLP Algorithm}\label{sect:CSGLP}

In this section, we present the CSGLP algorithm in detail. CSGLP aims to get better camera-invariant features and generalize classification performance well on the target domain, in which only unlabeled data are available.

\subsection{Problem Formulation}

Suppose that there is a source domain with $N_s$ labeled data $\mathcal{S}=\{x_{i}^{s}, y_{i}^{s}, c_{i}^{s}\}_{i=1}^{N_s}$ where $y_{i}^{s} \in \{1, 2, \cdots, M_s\}$ and $c_{i}^{s}$ is a $C_s$-dimensional one-hot vector indicating the camera label of sample $x_i^s$. Here $M_s$ is the number of pedestrian classes and $C_s$ is the number of camera classes in the source domain. And there is a target domain with $N_t$ unlabeled data $\mathcal{T}=\{x_{i}^{t}, c_{i}^{t}\}_{i=1}^{N_t}$, where $c_{i}^{t}$ is a $C_t$-dimensional one-hot vector indicating the camera label of $x_i^t$. We train a generator $G$ with $\mathcal{S}$ and $\mathcal{T}$, without pedestrian labels involved. Then we translate the labeled data in $\mathcal{S}$ to any camera style of the target domain $\mathcal{T}$, and denote the translated domain as $G(\mathcal{S})$. For simplicity, we denote one dataset as a \textit{domain}, and one camera as a \textit{sub-domain}.

As we mentioned above, training on the source domain $\mathcal{S}$ and then directly testing on the target domain $\mathcal{T}$ results in unsatisfied person re-ID performance. Note that this manner has nothing to do with either pedestrian category adaptation or camera style adaptation. In this paper, we adopt this method as the baseline and denote it as 'No adaptation' (NA).

CSGLP algorithm includes two modules. The first module constructs a generative framework to capture camera variations from the target domain to the source domain. The second module aims to further fine-tune the model by using the KNN-based soft-labeling method, which helps to mine intrinsic structure of the unlabeled target domain.

\subsection{Camera Styles Generation}\label{sect:camera-style}
Based on the GAN framework, image-to-image translation methods are developed to achieve data distribution translation and show impressive performance\cite{zhu2017unpaired, kim2017learning, yi2017dualgan, Choi01StarGAN}. Particularly, SPGAN~\cite{Deng01Image} and PTGAN~\cite{wei2017person} propose image-to-image translation methods individually to exploit the labeled source domain in the person re-identification field. The image generation strategy relieves the labeled data deficiency problem and achieves good performance on the target domain. However, both of them either transfer the global style of the source domain to the target domain, i.e., without camera style discrepancy, or train multiple CycleGANs to learn respective mappings between specific camera styles. Therefore, they are unable to capture the variations in the camera level and in an efficient way.

To address these problems and enhance camera style adaptation, we adopt a camera style generation framework to learn the mappings between different cameras (i.e., sub-domains) and multiple datasets ($\mathcal{T}\rightarrow\mathcal{S}$) simultaneously. In such a framework, we use only one generator to capture the between-domain and within-domain variations. For a given labeled image $(x_{i}^{s}, y_{i}^{s})$ in the source domain with camera label $c_{i}^{s}$, the generative framework should transfer it to another camera style $c_{}^{t}$ of the target domain. Here $c_{}^{t}$ is also one-hot vector, and the subscript is omitted to denote freely any camera style.

StarGAN~\cite{Choi01StarGAN} can realize this strategy by using a unique generator $G$ to learn the translation mappings among multiple domains, and a unique discriminator $D$ to fight against the generator $G$. Usually, $G$ takes an image, a domain (dataset) label and a sub-domain (camera) label as input, and outputs a translated image $G(x, c)$. $D$ is a mapping $D: x \to \{D_r(x), D_c(x)\}$ in which $D_r(x)$ is the probability of being a real image, playing the same role as the discriminator in the original GAN framework. $D_c(x)$ predicts all sub-domain (camera) labels in all domains and regularizes the generator $G$ to learn camera variations. In the perspective of network architecture, $G$ takes domain label and sub-domain label as additional channels, and puts their one-hot vectors into the network. In other words, in order to cooperate with domain and sub-domain labels simultaneously, it needs to constructs a mask vector. The whole label to be put into $G$ is defined as
\begin{equation*}
c = [c_{1}, c_{2}, mask],
\end{equation*}
in which $c_1$ and $c_2$ are one-hot vectors or zero vectors indicating the sub-domain label, while $mask$ is a one-hot vector indicating the domain label. More specifically, if the first element of $mask$ is 1, the sub-domain label $c_1$ is a one-hot vector while $c_2$ is a zero vector; Conversely, if the first element of $mask$ is 0, the formulation of $c_1$ and $c_2$ is interchanged.

As usual GANs, we have an adversarial loss to realize adversarial training, in which $D$ is trained to distinguish images from real images and generated images, and $G$ is trained to generate images that $D$ cannot distinguish correctly, i.e.,
\begin{equation}
\label{eq:gp}
\begin{split}
\mathop{\min}_{G} \mathop{\max}_{D} \mathcal{L}_{adv} = & \mathds{E}_{x}[\log D_r(x)] - \mathds{E}_{x, c}[\log (D_r(G(x, c)))]. \\
\end{split}
\end{equation}

To ensure that the generator distinguishes images from different cameras and learns the camera variations, the discriminator $D$ is trained on real images to capture the camera variations, and then $D$ drives the generator $G$ to generate target style-based images. Overall, a domain-oriented classification loss for $D$ and $G$ is considered. Training $D$ is to minimize
\begin{equation}
\mathcal{L}_c^{d} = \mathds{E}_{x, c'}[-\log D_c(c'|x)],
\end{equation}
and training $G$ is to minimize
\begin{equation}
\mathcal{L}_c^{g} = \mathds{E}_{x, c}[-\log D_c(c|G(x,c)),
\end{equation}
in which $c'$ is the true sub-domain label of input $x$, and $c$ is a random label to make $G$ learn the mapping well.

Finally, we need to encourage pedestrian information of the source domain to be preserved during the translation process, and only change the camera styles. Thus a reconstruction-based loss, which is adapted from a cycle consistency criterion~\cite{zhu2017unpaired, kim2017learning}, is applied to the generator as follows:
\begin{equation}
\mathcal{L}_{rec} = \mathds{E}_{x, c, c'}[||x-G(G(x,c),c')||_{1}].
\end{equation}
Above of all, the objective function with respect to $D$ is
\begin{equation}
\mathop{\max}_{D} \mathcal{L}_{D} = \mathcal{L}_{adv} - \lambda_c\mathcal{L}_c^{d},
\end{equation}
and the objective function to optimize $G$ is
\begin{align}
\label{eq:cls and rec}
\mathop{\min}_{G} \mathcal{L}_{G} = \mathcal{L}_{adv} + \lambda_c\mathcal{L}_c^{g} + \lambda_{rec}\mathcal{L}_{rec}.
\end{align}
Here, $\lambda_c$ and $\lambda_{rec}$ are positive regularization parameters.

After training StarGAN, for the labeled data from the source domain, we keep the pedestrian labels unchanged and randomly choose one of the target camera labels to translate those pedestrian images. We finally get a translated source domain $G(\mathcal{S})=\{G(x_{i}^{s}), y_{i}^{s}, c_{i}^{t}\}_{i=1}^{N_s}$ with target camera labels $c_{i}^{t}$. The translated domain will have approximate variations, like similar lighting conditions, resolutions, backgrounds, etc., with the target domain. In other words, the generator $G$ constructs an appropriate transition space for improving feature matching. Then the soft-labeling method can be used directly to make label prediction/propagation for the samples in the target domain.

It is worth noting that the Local Max Pooling (LMP) method~\cite{Deng01Image} shows effectiveness in improving classification performance on the target domain, by mitigating the influence of noise. Such procedure provides a finer partition by locally dividing the output of Conv5 in ResNet50 into several small parts and perform local pooling on each part. This leads to higher discriminative descriptors, and thus improves the re-ID accuracy. The same strategy is also adopted by our work, and the performance boosting will be independently shown in the following sections.

\subsection{Cross-domain Label Propagation}
We now have the translated source domain $G(\mathcal{S})$ and the unlabeled target domain $\mathcal{T}$ on hand. The person re-ID model can be trained directly on $G(\mathcal{S})$, and then fine-tuned on $\mathcal{T}$. However, the non-overlapping-label challenge that subjects in different domains are totally different emerges. In this subsection, we propose a KNN-based method to generate soft labels for the unlabeled data in $\mathcal{T}$ according to their distances to the translated data in $G(\mathcal{S})$.

Our motivation comes from the fact that $G(\mathcal{S})$ has similar data distribution with $\mathcal{T}$, thus they share some characteristics like similar lighting conditions, resolutions, backgrounds, etc. For the model trained on $G(\mathcal{S})$ which can be used to learn the target camera-invariant features, these translated labeled identities lie in a discriminative subspace and images of the same identity are close to each other. Notice that the non-parametric KNN is sensitive to the local structure of the data. This facilitates us to mine auxiliary information from the relationship between $G(\mathcal{S})$ and $\mathcal{T}$, and make it possible to learn from the target domain $\mathcal{T}$.

We first use current model to extract features of the unlabeled data, and then generate soft label for each unlabeled sample based on its feature distances with those translated labeled data. Specifically, we denote the person re-ID network mapping as $f$. Suppose $G(\mathcal{S})$ contains $N_s$ pedestrian images and their labels, i.e., $G(\mathcal{S}) = \{f(G(x_i^s)), y_i^s\}_{i=1}^{N_s} $, $y_i^s\in \{1,2,\dots,M_s\}$, $M_s$ is the number of classes in the source domain. For any unlabeled feature representation $f(x_j^t)$ in the target domain, we choose the $K$ nearest images from the source domain according to their distances, denoted as $\{f(G(x_{i_k}^s)), y_{i_k}^s\}_{k=1}^{K}$. Then similar to~\cite{salakhutdinov2007learning}, the probability that the point $x_j^t$ selects one of its neighborhoods, $G(x_{i_k}^s)$, can be formulated as
\begin{equation}
P_{j, k} = \frac{\exp(-\lambda \Vert f(x_{j}^{t})-f(G(x_{i_{k}}^{s})) \Vert_{}^{2})}{\sum_{k}\exp(-\lambda \Vert f(x_{j}^{t})-f(G(x_{i_{k}}^{s})) \Vert_{}^{2})},
\end{equation}
and the probability that the point $x_{j}^{t}$ belongs to class $p \in \{1, 2, \dots, M_s\}$ can be estimated via accumulating the probabilities over class $p$ included in $K$ nearest neighborhoods:
\begin{equation}
\label{eq:assign probability}
y_{jp}\!\triangleq\!P(y_j^t\!=\!p)\!=\!\sum_{k: y_{i_k}^s=p} P_{j, k}.
\end{equation}


\begin{figure}[htbp]
	\centering{{\includegraphics[width=0.48\textwidth]{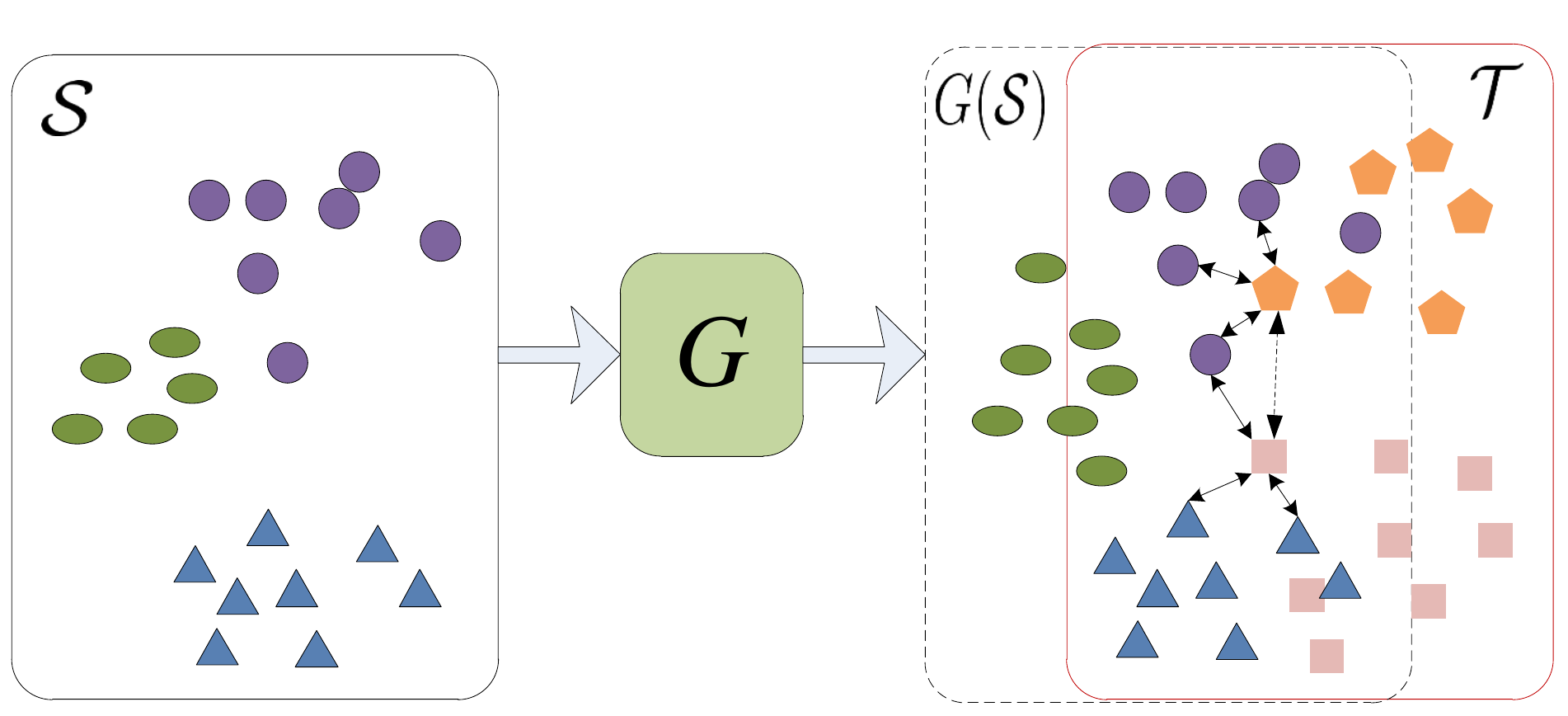}}\caption{Sketch map of the soft labeling method. $G$ is the camera style generator/translator from the target domain $\mathcal{T}$ to the source domain $\mathcal{S}$. Thus the soft labeling method is performed in the union $G(\mathcal{S})$ and $\mathcal{T}$. The soft labels reflect partially the neighborhood relationship between orange pentagon and pink square. Better viewed in color.}\label{fig:soft labeling}} 
\end{figure}

For an unlabeled sample $x_{j}^{t}$ in the target domain $\mathcal{T}$, we choose its $K$ nearest neighbors from $G(\mathcal{S})$ and compute its soft label vector. A toy example illustrating our motivation of soft labeling method is shown in Fig.~\ref{fig:soft labeling}. In this figure, we show the case of $K=3$. $K$ nearest neighbors of the orange pentagon are all purple circles, so its soft label vector is one-hot. The pink square is close to two blue triangles and one purple circle, thus it will be assigned to blue triangle class with a large probability, and to purple circle class with a small probability. Both the orange pentagon and the pink square we considered will be assigned to green ellipse with zero probability. Then we can infer that orange pentagon and pink square come from different classes.

Two important parameters are introduced to the definition of soft labels $P(y_j^t=p)$, as shown in Eq.~\eqref{eq:assign probability}. (1) A proper $K$ value is important to generate good soft labels. Since our goal is to capture the intrinsic locality structure of images in $\mathcal{T}$, based on their $K$ nearest neighborhoods in the translated source domain $G(\mathcal{S})$, a small $K$ value will be inadequate to reveal the local structure. In contrary, a large $K$ value will lead to a situation that the identities having more images will dominate the soft labels by numerical accumulation. (2) The hyper-parameter $\lambda$ is used to control importance of the nearest neighborhoods according to the distances. A large $\lambda$ value corresponds to a large probability to those identities who are very close to the unlabeled samples, and it enforces the model to concentrate on more informative samples.

After obtaining the soft labels for images in the target domain $\mathcal{T}$, we fine-tune the re-ID model to enforce the softmax distribution of the target data to approximate the soft labels by using the cross-entropy loss, namely,
\begin{equation}
\label{eq:cross-entropy loss}
\mathcal{L}_{cross-entropy} = - \frac{1}{N_t}\sum_{i=1}^{N_t} \sum_{p=1}^{M_s}y_{jp}\log \hat{y}_{jp},
\end{equation}
where $\hat{y}_{jp}$ is the output of softmax layer. The model trained on such \textit{labeled} target data is expect to generalize well on the target domain.

\section{Experiments}\label{sect:experiments}
In this section, we evaluate the CSGLP algorithm on two benchmark datasets, and then compare the results with some state-of-the-art methods.

\subsection{Datasets and Implementation Settings}
We choose two large-scale person re-ID datasets for experiments to evaluate the CSGLP algorithm, i.e., Market-1501\cite{zheng2015scalable} and DukeMTMC-reID\cite{ristani2016performance}.

Market-1501 dataset consists of 1,501 identities from 6 cameras. There are 751 identities with 12,936 training images for training, while the other 750 identities with 19,732 gallery images for testing. Each identity is captured by at most 6 cameras. All the bounding boxes are produced by DPM~\cite{felzenszwalb2008discriminatively} rather than manual annotation.

DukeMTMC-reID is a subset of DukeMTMC~\cite{ristani2016performance}. It contains 34,183 image boxes of 1,404 identities from 8 cameras: 702 identities are used for training and the remaining 702 for testing. There are 2,228 queries and 17,661 gallery images.

In the following we describe implementation detail of our experiments. We adopt the architecture of StarGAN to achieve the style adaptation between different cameras of different datasets, except that we use $(256, 128)$ as input size. It does not affect the generator, but some modifications have to be made for the discriminator to make it accept $(256, 128)$ as input size. The output Layer $D_r$ will be modified from $\rm CONV\textrm{-}(N1, K3x3, S1, P1)$ to $\rm CONV\textrm{-}(N1, K4 x 2, S1, P0)$, and the output Layer $D_c$ will be modified as $\rm CONV\textrm{-}(Nc\_dim, K4 x 2, S1, P0)$. In order to make the training process more stable, the Wasserstein GAN objective with gradient penalty~\cite{arjovsky2017wasserstein, gulrajani2017improved} is exploited.

We use the Adam \cite{kingma2014adam} optimizer with $\beta_1 = 0.5$ and $\beta_2 = 0.999$ to train StarGAN. The batch size is set to 16. We perform one generator update after every five discriminator updates as in \cite{gulrajani2017improved} and train the model 50,000 iterations.

In the feature learning stage, we choose the ResNet-50 \cite{He2016ResNet} and GoogLeNet \cite{Szegedy2015going} as person re-ID baseline models, respectively. Results on the Market1501 and DukeMTMC datasets are shown in Table~\ref{table:market1501} and Table~\ref{table:dukemtmc}, respectively. During training StarGAN, cameras labels and dataset labels are needed, while person labels are not.

For evaluation protocols, we report the mean average precision (mAP) and the rank-1, 5, 10 accuracies. All experiments use single query~\cite{Fan01Unsupervised} throughout this paper.

\subsection{Comparisons with State-of-the-art Methods}
In this section, we present quantitative results of our CSGLP algorithm. We compare CSGLP with several state-of-the-art approaches, e.g., Bow~\cite{zheng2015scalable}, LOMO~\cite{liao2015person}, UMDL~\cite{peng2016unsupervised}, PUL~\cite{Fan01Unsupervised}, CAMEL~\cite{yu2017cross}, SPGAN~\cite{Deng01Image} and PTGAN~\cite{wei2017person}. Although \cite{ye2017dynamic} and \cite{liu2017stepwise} also involve the label estimation stage, they are based on the tracklet data~\cite{liu2017stepwise} while our CSGLP algorithm is directly based on images. To make fair comparisons, we use our 2048 dim ResNet-50 features for implementing CAMEL.

Since SPGAN uses the ResNet-50 network structure and PTGAN uses the GoogLeNet structure for person re-ID, the CSGLP method uses both ResNet-50 and GoogLeNet as baseline network structures integrating the new loss functions for fair comparison. The re-ID accuracies of both configurations will be presented in the subsequent parts.

\begin{table}[htb]
	\caption{Performance comparison on Market1501.}
	\centering
	\renewcommand{\tabcolsep}{0.5pc} 
	\renewcommand{\arraystretch}{1.2} 
	\label{table:market1501}
	\scalebox{0.85}{
	\begin{tabular}{c|cccc}
		\hline
		Methods & rank-1 & rank-5 & rank-10 & mAP \\
		\hline
		Bow \cite{zheng2015scalable}           & 35.8 & 52.4 & 60.3 & 14.8 \\
		LOMO \cite{liao2015person}             & 27.2 & 41.6 & 49.1 & 8.0  \\
		\hline
		UMDL\cite{peng2016unsupervised}        & 34.5 & 52.6 & 59.6 & 12.4 \\
		NA (ResNet-50)                     & 42.8 & 60.1 & 68.3 & 19.0 \\
		PUL (ResNet-50)\cite{Fan01Unsupervised}       & 45.5 & 60.7 & 66.7 & 20.5 \\
		CAMEL (ResNet-50) \cite{yu2017cross}     & 50.4 & 67.8 & 74.1 & 21.8 \\
		\hline
		CycleGAN (ResNet-50) \cite{Deng01Image}       & 45.6 & 63.8 & 71.3 & 19.1 \\
		SPGAN (ResNet-50) \cite{Deng01Image}          & 51.5 & 70.1 & 76.8 & 22.8 \\
		HHL (ResNet-50)~\cite{Zhong2018retri}         & \textbf{60.3} & 77.3 & \textbf{84.0} & \textbf{31.4} \\
		StarGAN (ResNet-50)                           & 55.6 & 74.7 & 80.6 & 27.4 \\
		CSGLP (ResNet-50)                             & 59.2 & 76.2 & 83.2 & 31.1 \\
		\hline
		NA (GoogLeNet)          & 35.9 & 55.6 & 63.3 & 16.1 \\
		PTGAN(GoogLeNet)\cite{wei2017person}& 38.6 & -  & 66.1 & -  \\
		StarGAN(GoogLeNet)                & 51.3 & 73.0 & 80.2 & 25.8 \\
		CSGLP(GoogLeNet)                  & 58.8 & \textbf{77.6} & 83.2 & 30.9 \\
		\hline
	\end{tabular}}
\end{table}

\subsubsection{\textbf{Performance on Market1501}}
When the re-ID models are evaluated on Market1501, it is used as the target domain, and DukeMTMC-reID is used as the source domain.

\begin{figure}[htbp]
\centering{{\includegraphics[width=0.42\textwidth]{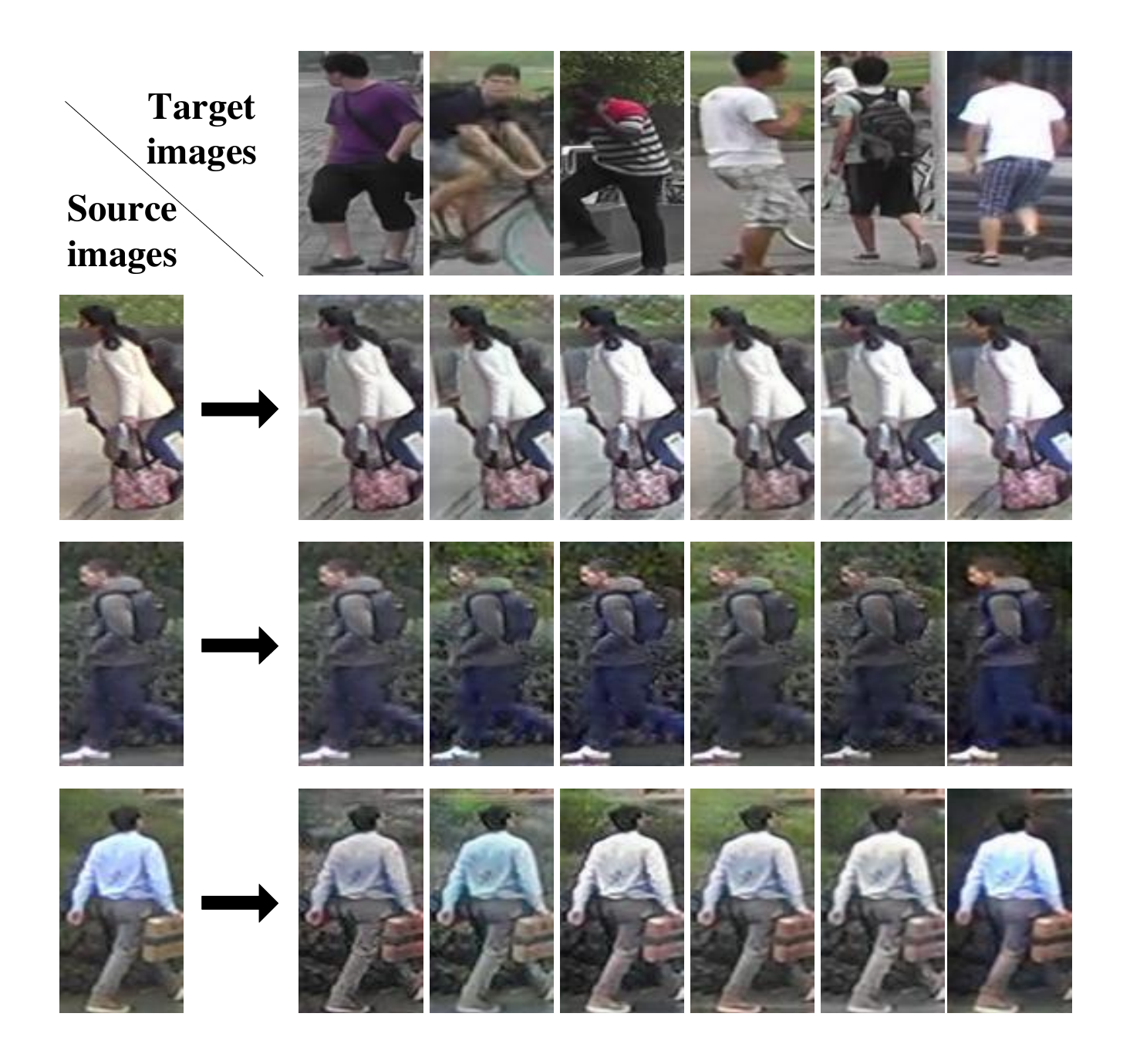}}\caption{Translated samples with DukeMTMC-reID as the source domain. The first row shows original images from Market1501. In the rows 2-4, column 1 shows images from DukeMTMC-reID, columns 2-7 are images translated to six camera styles of Market1501. Better viewed in color.}\label{fig:translated_results_d2m}} 
\end{figure}

We use StarGAN to transfer the camera styles of Market1501 to images of DukeMTMC-reID, and show some translated samples in Fig.~\ref{fig:translated_results_d2m}. The three images picked from the source domain have different actions and poses, while images of the target domain have different backgrounds and lighting conditions. In particular, the second and fourth target images have brighter background than others. We see that the main content and pedestrian identity of the translated images do not change, but the background and style are adapted. In other words, the target camera styles are captured by the generator, and the target camera variations are achieved.

We use these translated images, as shown in Fig.~\ref{fig:translated_results_d2m}, to train the baseline classification model with ResNet-50, and present the re-ID results in Table~\ref{table:market1501}. We denote this simple combination as 'StarGAN (ResNet-50)' for convenience. This method achieves 55.6\% rank-1 accuracy, and results in improvements around 20\% compared with traditional methods BOW and LOMO. It also gets 21.1\%, 10.1\% and 5.2\% improvements compared with UMDL, PUL and CAMEL, respectively. The performance enhancement demonstrates camera-invariant features can be learned effectively with the help of generator network. And we can see 12.8\% improvement in rank-1 accuracy testing on Market1501, compared with NA method, while 10.0\% and 4.1\% improvement with CycleGAN and SPGAN, respectively. When GoogLeNet is used as the base model, we see 15.4\% and 12.7\% improvement comparing with NA method and PTGAN, respectively. The gaps between StarGAN and other translation frameworks such as SPGAN and PTGAN are mainly due to that it captures more specific camera variations, thus the person re-ID model can learn camera-invariant features better.

After getting the soft labels, we fine-tune the classification model to learn from the unlabeled target domain. Results are presented in Table~\ref{table:market1501}. This method further improves performance of the modified ResNet-50 model and GoogLeNet model with 3.6\%, 7.5\% increase in the rank-1 accuracy, comparing with the StarGAN model which is trained on the translated DukeMTMC-reID data without soft-labeling.

When comparing with related state-of-the-art methods, we observe that CSGLP outperforms other methods by a large margin except for HHL. For example, we have 23.4\% and 32.0\% improvements compared with hand-crafted features methods BOW\cite{zheng2015scalable} and LOMO\cite{liao2015person}. The rank-1 accuracy of PUL and CAMEL is 30.0\% and 38.6\%, respectively. When comparing with the state-of-the-art unsupervised learning methods, our re-ID accuracy also have comparative results. For example, we see 7.7\% improvement on Market1501 compared with SPGAN. We observe that the results of HHL are slightly better than those of CSGLP, except for the rank-5 accuracy with the GoogLeNet architecture. In fact, HHL aims to extract camera-specific information from the target domain for feature fusion and enhancement, thus, it gets better performance on the Market1501 dataset.

\begin{table}[htb]
	\caption{Performance comparison on DukeMTMC-reID.}
	\centering
	\renewcommand{\tabcolsep}{0.5pc} 
	\renewcommand{\arraystretch}{1.2} 
	\label{table:dukemtmc}
	\scalebox{0.83}{
	\begin{tabular}{c|cccc}
		\hline
		Methods & rank-1 & rank-5 & rank-10 & mAP \\
		\hline
		Bow \cite{zheng2015scalable}           & 17.1 & 28.8 & 34.9 & 8.3  \\
		LOMO \cite{liao2015person}             & 12.3 & 21.3 & 26.6 & 4.8  \\
		\hline
		UMDL~\cite{peng2016unsupervised}        & 18.5 & 31.4 & 37.6 & 7.3  \\
		NA (ResNet-50)                     & 28.1 & 44.9 & 51.8 & 15.8 \\
		PUL (ResNet-50) \cite{Fan01Unsupervised}      & 30.0 & 43.4 & 48.5 & 16.4 \\
		CAMEL (ResNet-50) \cite{yu2017cross}     & 38.6 & 56.1 & 63.3 & 21.4 \\
		\hline
		CycleGAN (ResNet-50)~\cite{Deng01Image}        & 38.1 & 54.4 & 60.5 & 19.6 \\
		SPGAN (ResNet-50)~\cite{Deng01Image}           & 41.1 & 56.6 & 63.0 & 22.3 \\
		HHL (ResNet-50)~\cite{Zhong2018retri}         & 44.7 & 61 & 66.3 & 25.5 \\
		StarGAN (ResNet-50)                           & 42.9 & 59.1 & 65.7 & 24.1 \\
		CSGLP (ResNet-50)                              & \textbf{47.8} & \textbf{62.3} & \textbf{68.3} & \textbf{27.1} \\
		\hline
		NA (GoogLeNet)           & 19.5 & 32.4 & 38.6 & 8.8  \\
		PTGAN (GoogLeNet)~\cite{wei2017person}& 27.4 & -    & 50.7 & -    \\
		StarGAN (GoogLeNet)               & 36.8 & 52.6 & 59.9 & 20.1 \\
		CSGLP (GoogLeNet)                   & 39.0 & 56.2 & 63.4 & 20.8 \\
		\hline
	\end{tabular}}
\end{table}

\subsubsection{\textbf{Performance on DukeMTMC-reID}}
In this section, we conduct experiments on DukeMTMC-reID when Market1501 is used as the source domain.

\begin{figure}[htbp]
	\centering{{\includegraphics[width=0.46\textwidth]{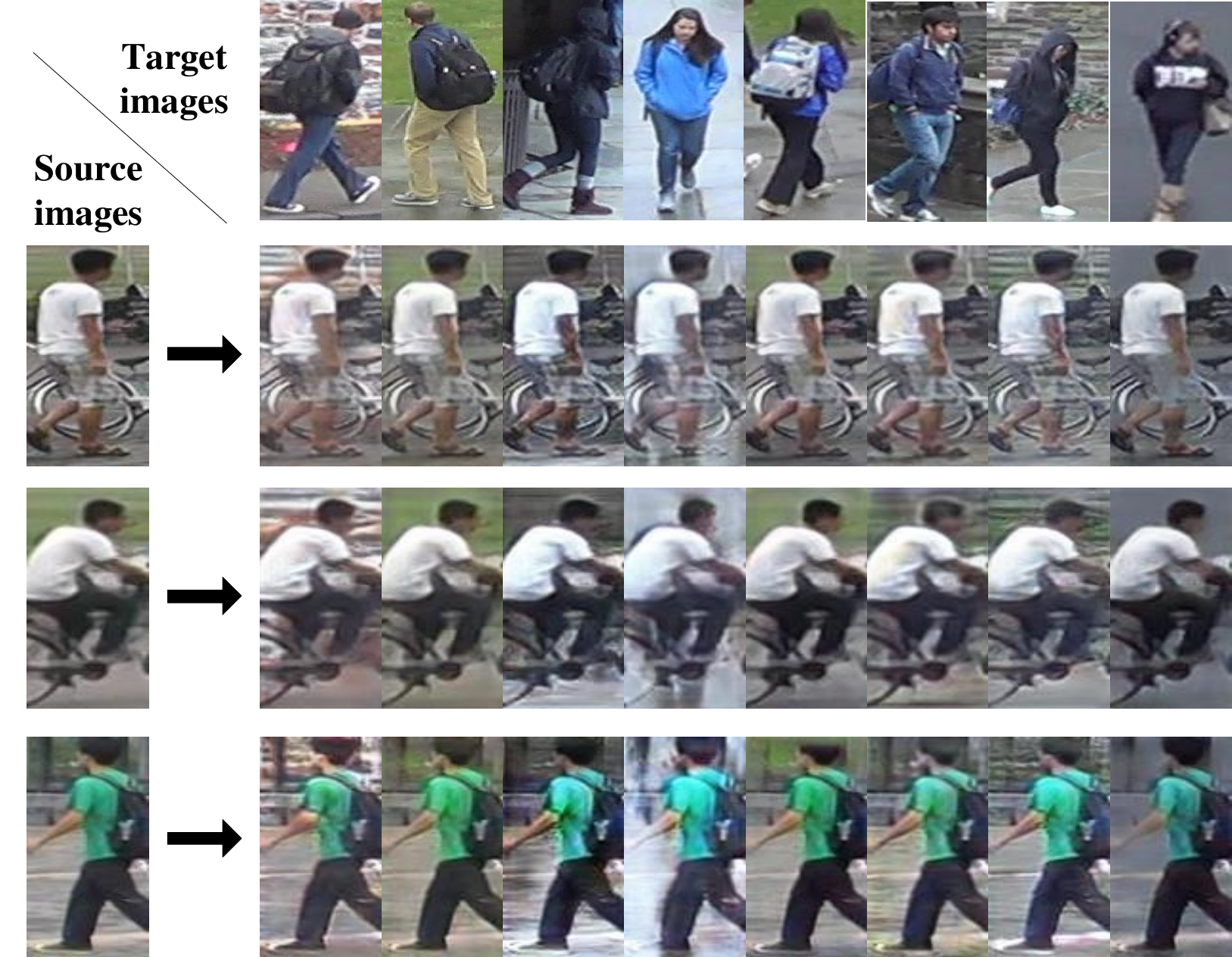}}\caption{Translated samples with Market1501 as the source domain. The first row shows original images from DukeMTMC-reID. In the rows 2-4, column 1 show images from Market1501, columns 2-9 are images translated to eight camera styles of DukeMTMC-reID. Better viewed in color.}\label{fig:translated_results_m2d}} 
\end{figure}

At the camera style adaptation stage, we get the translated Market1501 dataset, and show some translated samples in Fig.~\ref{fig:translated_results_m2d}. Three images picked from the source domain have different behaviors and postures, while images of the target domain have different backgrounds and lighting. We see that the translated source images do incorporate the camera styles of the target domain. Taking images in the fourth column for example, the bright illumination style is transferred well to the source images. In other words, the target camera styles are captured by the generator, and the target camera variations are achieved.

We show the re-ID results for DukeMTMC-reID in Table~\ref{table:dukemtmc}. We see that even trained with the ResNet-50 network structure and the softmax classification loss function, the deep network without domain adaptation module, i.e., NA, obtains the rank-1, rank-5, and rank-10 re-ID accuracy 28.1\%, 44.9\%, and 51.8\%, respectively. Compared with the ResNet-50 structure, the GoogLeNet structure obtains lower accuracies. However, after using StarGAN to perform the camera style translation from the target domain to the source domain, the re-ID accuracy obtained by the style-transferred images is increased to 42.9\%, 59.1\% and 65.7\%, respectively. Furthermore, CSGLP fine-tunes the classification model by using the soft labels, and increases the results to 47.8\%, 62.3\% and 68.3\%, respectively. The performance improvement is significant.

For other methods based on camera style transfer learning, such as CycleGAN, SPGAN and HHL, we can see that the rank-1 accuracy is 38.1\%, 41.1\% and 44.7\% respectively. Thus, CSGLP outperforms these methods. The remaining evaluation criteria including rank-5, rank-10 and mAP results also indicate the superiority of CSGLP. Besides, in terms of the results based on the GoogLeNet network structure, we see that CSGLP also significantly outperforms other methods.

\subsubsection{\textbf{Ablation study for LMP}}

As we stated in Section~\ref{sect:camera-style}, LMP provides us a finer partition by locally dividing the output of Conv5 in ResNet50 into $P$ parts and perform local pooling on each part. It has been empirically validated by SPGAN~\cite{Deng01Image} that direct application of LMP into the network can lead to higher discriminative descriptors, and thus improve the re-ID accuracy. In our work, since our input size of images is 256$\times$128, before global average pooling in ResNet-50 is the output with size 2048$\times$9$\times$5. So we adopt $P=9$ in LMP to improve the results on both Market1501 and DukeMTMC-reID. We show the final re-ID accuracies in Table \ref{table:results1}.

\begin{table}[htb]
	\caption{Applying LMP for further improving the classification performance.}
	\centering
	\renewcommand{\tabcolsep}{0.3pc} 
	\renewcommand{\arraystretch}{1.2} 
	\label{table:results1}
	\scalebox{0.8}{
	\begin{tabular}{c|c|cccc}
		\hline
		Dataset &Methods & rank-1 & rank-5 & rank-10 & mAP \\
		\hline
		\multirow{2}*{Market1501} & SPGAN+LMP \cite{Deng01Image}  & 58.1 & 76.0 & 82.7 & 26.9   \\
		& CSGLP+LMP & \textbf{63.7} & \textbf{79.8} & \textbf{85.2} & \textbf{33.9} \\
		\hline
		\multirow{2}*{DukeMTMC-reID} & SPGAN+LMP \cite{Deng01Image}  & 46.9 & 62.6 & 68.5 & 26.4   \\
		& CSGLP+LMP & \textbf{51.2} & \textbf{65.8} & \textbf{71.2} & \textbf{30.3} \\
		\hline
	\end{tabular}}
\end{table}

For the Market1501 dataset, the rank-1 accuracy of CSGLP without LMP is 59.2\%, however, it increases to 63.7\% by applying LMP into the network learning procedure. It indicates that the local max pooling approach indeed improves the domain adaptation and final classification performance. Besides, we can see that CSGLP+LMP outperforms both SPGAN and SPGAN+LMP, which achieves the accuracy 51.5\% and 58.1\%, respectively.

For the DukeMTMC-reID dataset, the rank-1 accuracy of CSGLP+LMP is 51.2\%, which is obviously higher than that of CSGLP (47.8\%). Besides, we can see that CSGLP+LMP outperforms both SPGAN and SPGAN+LMP, which achieves the accuracy 41.1\% and 46.9\%, respectively. It indicates again that the LMP module indeed improves the domain adaptation and final classification performance.

\section{Conclusion}\label{sect:conclusion}

In this paper, we consider an unsupervised domain adaptation problem for person re-ID. It has several difficulties caused by a large domain discrepancy, which can be brought by either large camera variations or the non-overlapping-label challenge. To tackle these problems and improve effectively the classification performance, we exploit StarGAN to learn the mappings between different camera views of multiple domains. This image-to-image translation framework can capture the camera variations with only one generator and translate the labeled images in $\mathcal{S}$ to those cameras of the target domain $\mathcal{T}$. Besides, we propose a KNN-based method to predict soft labels for the unlabeled data of the target domain, based on their similar camera styles with the translated source domain. The CSGLP algorithm can alleviate heavy afford of labeling identities across ocean of cameras, and obtain state-of-the-art person re-ID results.

In the future, we will consider multi-source domain adaptation problem and more general application situations when more or larger datasets are available.


{\small
\bibliographystyle{IEEETrans}
\bibliography{ref_csglp}
}


\end{document}